\documentclass[letterpaper]{article} 
\usepackage{aaai24}  
\usepackage{times}  
\usepackage{helvet}  
\usepackage{courier}  
\usepackage[hyphens]{url}  
\usepackage{graphicx} 
\usepackage{natbib}  
\usepackage{caption} 
\usepackage{algorithm}
\usepackage{algorithmic}
\usepackage{amsmath}
\usepackage{amsthm}
\usepackage{amssymb}
\usepackage{dsfont}
\usepackage{algorithm}
\usepackage{algorithmic}
\usepackage{cleveref}
\usepackage{multirow}
\usepackage{multicol}
\usepackage{booktabs}
\usepackage{makecell}
\usepackage{subfig}
\usepackage{tabularx}

\usepackage{newfloat}
\usepackage{listings}
\DeclareCaptionStyle{ruled}{labelfont=normalfont,labelsep=colon,strut=off} 
\floatstyle{ruled}
\newfloat{listing}{tb}{lst}{}
\floatname{listing}{Listing}
\title{Noisy Correspondence Learning with Self-Reinforcing Errors Mitigation}
\author{
   Zhuohang Dang\textsuperscript{\rm 1}\thanks{This work was completed during his internship at SGIT AI Lab, State Grid Corporation of China.},\;  Minnan Luo\textsuperscript{\rm 1}\thanks{Corresponding author.},\; Chengyou Jia\textsuperscript{\rm 1},\; Guang Dai\textsuperscript{\rm 2,3},\; Xiaojun Chang\textsuperscript{\rm 4,6}, \\  Jingdong Wang\textsuperscript{\rm 5}
}
\affiliations{
    \textsuperscript{\rm 1}School of Computer Science and Technology, MOEKLINNS Laboratory, Xi'an Jiaotong University \\
    \textsuperscript{\rm 2}SGIT AI Lab \quad  \textsuperscript{\rm 3}State Grid Corporation of China  \;  \textsuperscript{\rm 4}University of Technology Sydney  \;  \textsuperscript{\rm 5}Baidu Inc   \\  \textsuperscript{\rm 6}Mohamed bin Zayed University of Artificial Intelligence \\
\tt\small \{dangzhuohang,cp3jia\}@stu.xjtu.edu.cn \ \tt\small minnluo@xjtu.edu.cn \\
\tt\small \{guang.gdai, cxj273\}@gmail.com \ \tt\small wangjingdong@outlook.com
}

\usepackage{bibentry}

\usepackage{xspace}
\makeatletter
\DeclareRobustCommand\onedot{\futurelet\@let@token\@onedot}
\def\@onedot{\ifx\@let@token.\else.\null\fi\xspace}
\def\eg{\emph{e.g}\onedot} 
\def\ie{\emph{i.e}\onedot}

\def\etc{\emph{etc}\onedot}

\makeatother

\newtheorem{theorem}{Theorem}
\begin{document}

\maketitle
\begin{abstract}
Cross-modal retrieval relies on well-matched large-scale datasets that are laborious in practice.
Recently, to alleviate expensive data collection, co-occurring pairs from the Internet are automatically harvested for training.
However, it inevitably includes mismatched pairs, \ie, noisy correspondences, undermining supervision reliability and degrading performance.
Current methods leverage deep neural networks' memorization effect to address noisy correspondences, which overconfidently focus on \emph{similarity-guided training with hard negatives} and suffer from self-reinforcing errors.
In light of above, we introduce a novel noisy correspondence learning framework, namely \textbf{S}elf-\textbf{R}einforcing \textbf{E}rrors \textbf{M}itigation (SREM).
Specifically, by viewing sample matching as classification tasks within the batch, we generate classification logits for the given sample.
Instead of a single similarity score, we refine sample filtration through energy uncertainty and estimate model's sensitivity of selected clean samples using swapped classification entropy, in view of the overall prediction distribution. 
Additionally, we propose cross-modal biased complementary learning to leverage negative matches overlooked in hard-negative training, further improving model optimization stability and curbing self-reinforcing errors.
Extensive experiments on challenging benchmarks affirm the efficacy and efficiency of SREM.
\end{abstract}

\section{Introduction}
Cross-modal matching aims to retrieve relevant samples across different modalities, which has become a focal research area due to the prevalence of multimedia data.
Contemporary methods achieve semantic alignment using modal-specific encoders \cite{diao2021similarity,li2021align}. 
They project data into a unified feature space, where matched data from different modalities are drawn together, while mismatched ones are pushed apart.
To alleviate the laborious collection of well-matched data, recent datasets \cite{sharma2018conceptual} automatically collect co-occurring sample pairs from the Internet for training.
However, they contain around $20\%$ mismatched pairs\cite{sharma2018conceptual,huang2021learning}, namely noisy correspondences.
Encouraging these mismatched pairs to be similar will significantly degrade the matching performance.
\setlength{\belowcaptionskip}{-7mm}
\begin{figure}[!t]
  \centering
  \subfloat[shows self-reinforcing errors in the training procedure of the state-of-the-art MSCN's \cite{han2023noisy} on Flickr30K with 60\% synthetic noise. As training progresses, noisy samples are gradually included and consequently degrade model performance.]{{\includegraphics[width=1\linewidth]{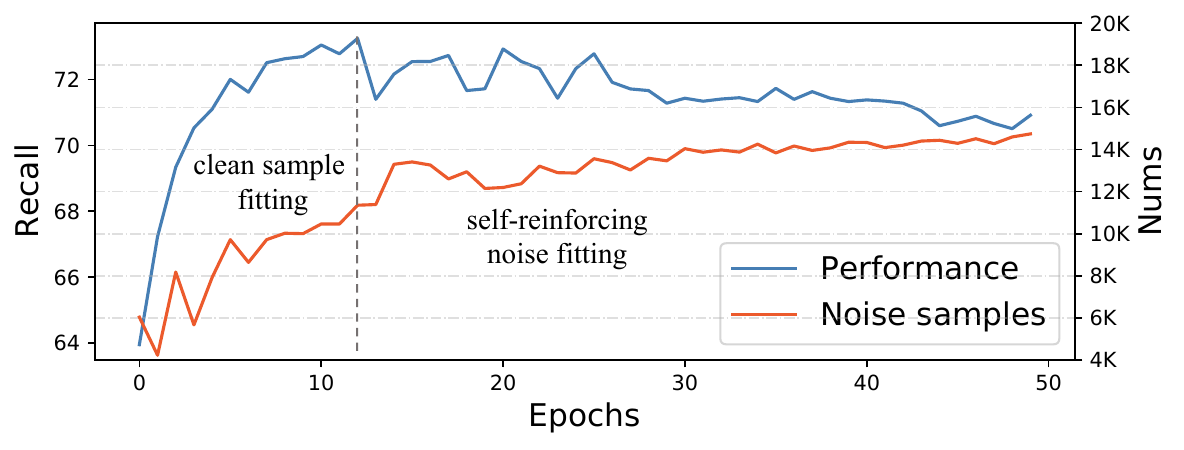}\label{fig: noise_accumulation}}}
  \\
  \subfloat[illustrates that hinge-based ranking loss, by solely focusing on query's positive and hard negative sample, yields sub-optimal results as the query inadvertently becomes closer to other negatives. ]{{\includegraphics[width=1\linewidth]{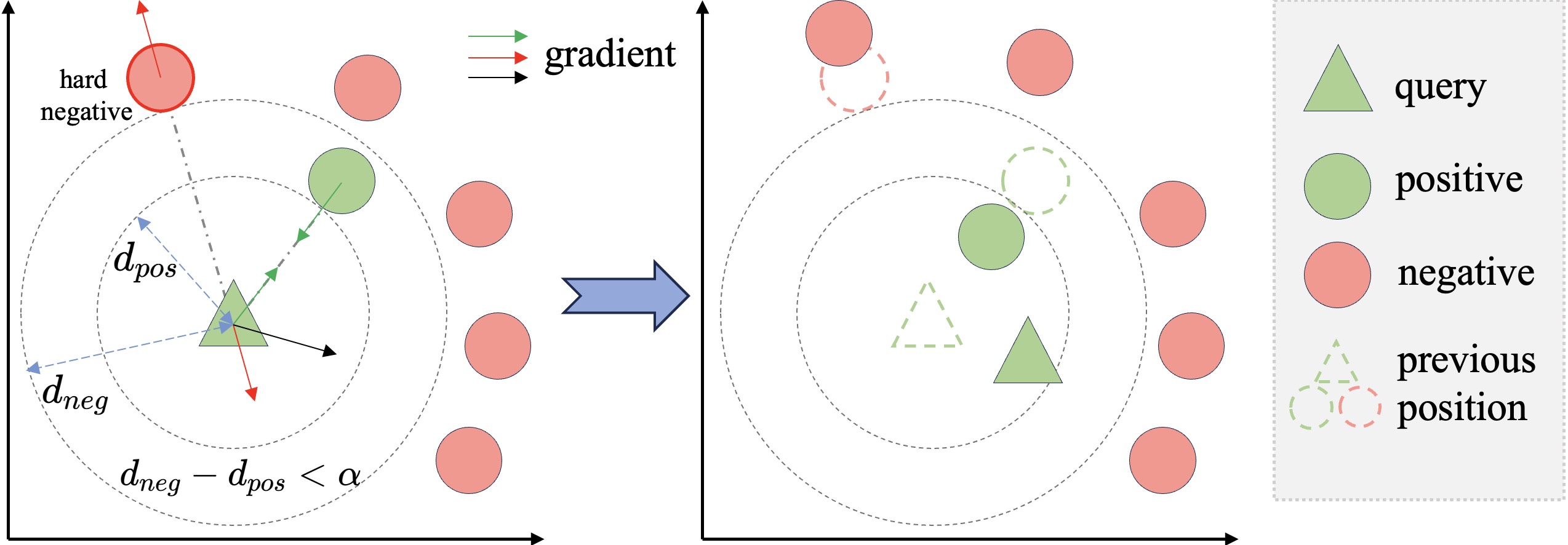}\label{fig: intro_2}}}
\caption{Drawback illustrations of similarity-guided training with hard negatives.}
\end{figure}
\setlength{\belowcaptionskip}{-0.0cm}

Recent advancements \cite{yang2023bicro,qin2022deep} have tackled noisy correspondences through deep neural network (DNN) memorization. This effect enables clean samples to exhibit higher similarities than noisy ones after the initial few epochs \cite{yao2020searching}.
Specifically, after warmup, these methods further refine similarity prediction with the following alternate steps:
1) Using similarity scores to identify clean samples. 
2) Deriving soft margins proportional to similarity scores for robust matching of selected clean samples. 
The soft margins are employed in a hinge-based ranking loss, where a larger margin intensifies the model's sensitivity towards differentiating the given sample from its negatives.
However, Figure 1(a) shows that such an approach is susceptible to self-reinforcing errors.
The primary vulnerability arises from the fact that the aforementioned two steps, clean sample selection and corresponding sensitivity estimation, rely heavily on the model's similarity prediction.
This leads to a critical issue where confident but incorrect similarity predictions are amplified during subsequent training, forming a loop of self-reinforcing errors \cite{chen2023two, yang2023bicro}.
Furthermore, hinge-based ranking loss solely focuses on the query's positive and hard negative sample, overlooking a wealth of negative information. 
Figure 1(b) shows that this narrow focus can result in suboptimal model optimization, potentially aggravating self-reinforcing errors.

In light of above, we propose a novel noisy correspondence learning framework, namely \textbf{S}elf-\textbf{R}einforcing \textbf{E}rrors \textbf{M}itigation (SREM). 
Specifically, SREM encompasses three core modules: 
\textbf{1)} 
We introduce a novel energy-guided sample filtration to complement conventional similarity-based sample filtration.
We first produce classification logits for a sample by viewing sample matching as a classification task within the batch.
We then use energy scores derived from classification logits to gauge the model's uncertainty during sample selection.
As a result, this strategy ensures the selected clean samples maintain both high similarity and low uncertainty, paving the way for more precise data division.
\textbf{2)} We propose a \textbf{S}wapped \textbf{G}radient \textbf{W}eighting (SGW) strategy.
SGW assesses the model's sensitivity towards individual samples by leveraging swapped classification entropy, ensuring robust matching of selected clean samples.
Samples with lower entropy suggest higher prediction confidence, thus the model should be more sensitive to them and let them contribute more to optimization \cite{iscen2019label}.
In contrast to a single similarity score, classification entropy considers the model's prediction distribution over both clean and negative samples, ensuring robustness.
\textbf{3)} 
We introduce a novel \textbf{C}ross-\textbf{M}odal \textbf{B}iased \textbf{C}omplementary \textbf{L}earning (CMBCL) objective for leveraging negative samples overlooked in the hinge-based ranking loss.
We perceive these overlooked negative matches as ``complementary labels" that essentially signal non-matching samples, guiding the model to distance positive samples from all negatives and thus circumventing potential self-reinforcing errors.

Extensive experiments highlight the substantial improvement achieved by SREM, surpassing state-of-the-arts by more than 1\% in average recall. 
Moreover, SREM also boasts a reduction in training time by more than 40\%, attesting its efficiency.
In addition to empirical validations, we theoretically prove the efficacy of CMBCL, as it converges to an optimal classifier equivalent to one trained with true labels.
We also highlight its generality, demonstrating that CMBCL encompasses the previously strong competitor RCL \cite{hu2023cross} as a special case.

\section{Related Work}
\subsection{Cross Modal Retrieval}
Cross-modal matching aims to project images and texts into a unified feature space where matched data from different modalities are similar while mismatched data are dissimilar.
This matching can be globally \cite{radford2021learning,chen2021learning}, by matching images and text from a comprehensive perspective, or locally \cite{diao2021similarity,lee2018stacked}, connecting specific regions within images to words in sentences for a more granular alignment.

Contrary to previous approaches that presuppose well-matched training data, the prohibitive collection costs have fostered the emergence of new paradigms like noisy correspondences, a prevalent issue in domains such as person re-id \cite{yang2022learning}, graph matching \cite{lin2023graph}, and multi-view learning \cite{yang2022robust,yang2021partially}.
Current methods in cross-modal matching \cite{yang2023bicro,han2023noisy,huang2021learning} primarily employ multi-step frameworks: They first estimate the distribution of instance-level loss/similarity across the entire dataset. 
Then they compute the posterior probability as the pseudo-label for each sample, which is further filtered by a threshold and clean samples are used for training.
To eliminate additional computation overhead caused by similarity distribution estimation, DECL \cite{qin2022deep} uses similarity with evidential learning to dynamically filter out noisy correspondences within each batch.
However, similarity-guided training in previous methods lead to self-reinforcing errors.
In contrast,  our SREM addresses overconfidence in similarity scores through overall prediction distributions, effectively mitigating such errors and notably enhancing performance.

\subsection{Complementary Label Learning}
Unlike conventional classification tasks, samples in complementary label learning (CLL) are assigned complementary labels that indicate classes they do not belong to.
To effectively use these weak supervisions, \cite{ishida2017learning,ishida2019complementary} assume the uniform distribution of complementary labels and prove an optimal classifier can be learned with mere complementary labels. 
Differently, some works \cite{yu2018learning,gao2021discriminative,xu2020generative} consider the unknown distribution of complementary labels. 
By estimating label transition probabilities, they inferred the distribution of complementary labels and subsequently refined them for training.
In noisy correspondence learning, RCL \cite{hu2023cross} extends CLL to introduce a novel contrastive learning framework that exclusively leverages negative information, mitigating the potential negative effects of mismatched samples.
However, the neglect of powerful positive supervision leads to suboptimal results for RCL.
On the contrary, beyond using positive supervision in  ranking loss, we additionally leverage the dissimilarity of negative samples to utilize negative information more effectively, therefore achieving a more robust training regime against noisy correspondence.

\section{Methodology}
\subsection{Problem Definition}
In line with previous works, we use image-text retrieval as a proxy task to explore the noisy correspondence in cross-modal matching, which consists of two sub-tasks: image-to-text (i2t) and text-to-image (t2i) retrieval.
Typically, we are provided with a training dataset $\mathcal{D}=\{(I_i,T_i,m_i)\}_{i=1}^N$, where $N$ is the data size and $(I_i,T_i,m_i)$ is the $i$-th image-text pair $(I_i,T_i)$ with label $m_i\in\{0,1\}$ indicating whether they are matched.
In noisy correspondence, an unknown portion of pairs in $\mathcal{D}$ is mismatched, \ie, the image and text are not matched but with matched labels.

\subsection{Model Overview}
\begin{figure}[!t]
  \centering
  \includegraphics[width=1\linewidth]{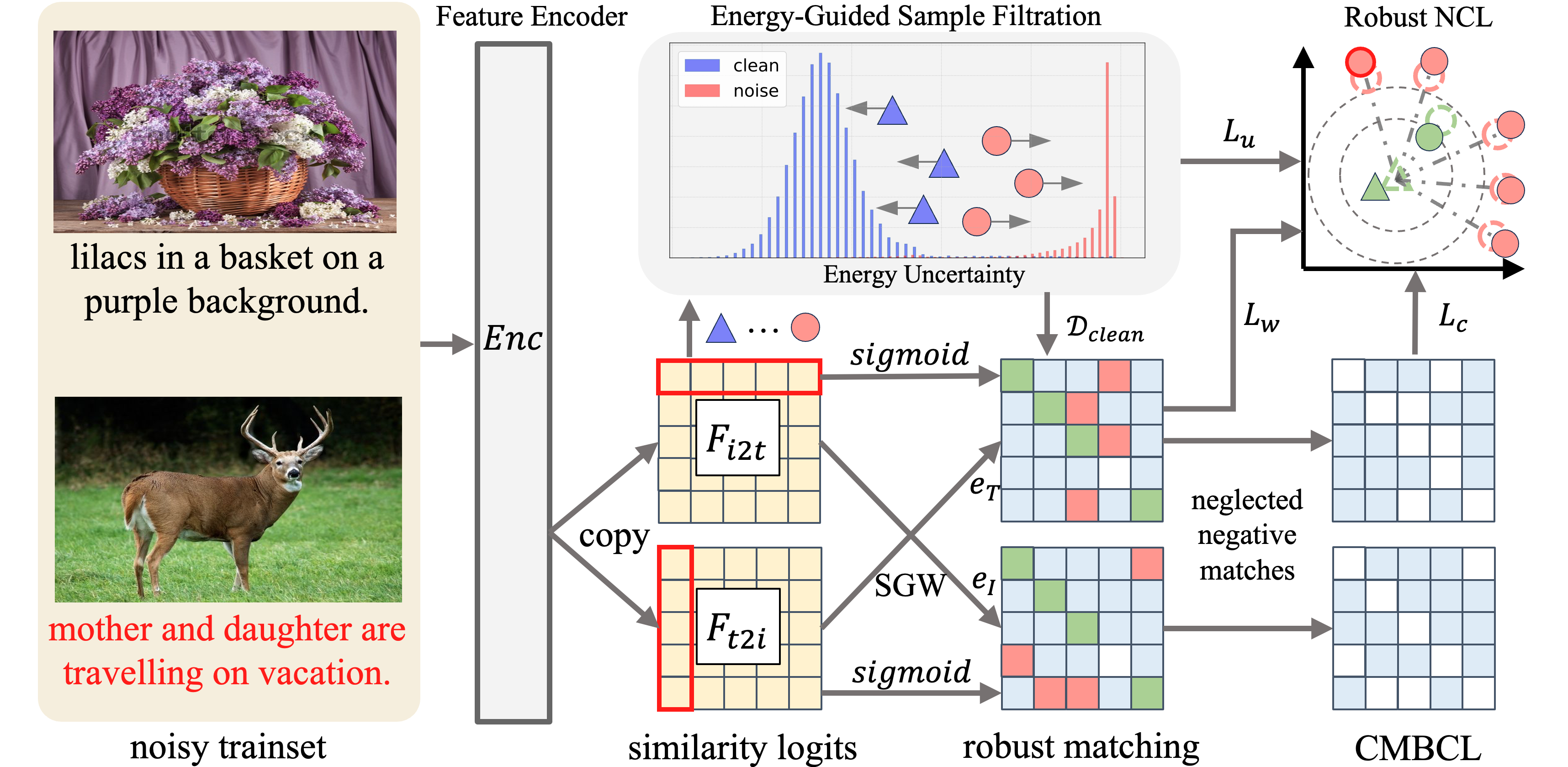}
\caption{Illustration of the proposed SREM.}
\label{fig: overview}
\end{figure}

In this section, we present our SREM in detail, whose overview is shown in \Cref{fig: overview}. 
For simplicity, we take image-to-text retrieval as a showcase to introduce the pipeline of SREM, while text-to-image retrieval is conducted in a symmetric manner.
Initially, the feature encoder generates similarity logits from the input pair. 
Then, we employ three elaborately-designed modules to mitigate the self-reinforcing errors during training.
Given the disparities in prediction distribution, we utilize energy uncertainty to segregate clean samples, denoted as $\mathcal{D}_{clean}$, from noisy correspondences, $\mathcal{D}_{noisy}$.
To enhance SREM's robustness, we introduce the swapped gradient weighting and cross-modal biased complementary learning framework.
The former proposes a gradient-rescaled ranking loss $L_w$, while the latter effectively leverages the overlooked negative matches in $L_w$ as complementary labels.
We will detail each component and corresponding optimization objective in what follows.

\setlength{\belowcaptionskip}{-6mm}
\setlength{\belowcaptionskip}{-0mm}
\subsection{Feature Encoder}
Initially, the feature encoder projects both visual and textual data into a unified feature space using model-specific encoders $f$ and $g$, respectively.
Within the unified feature space, a function $h$ computes the similarity logit as $F_{ij} = h(f(I_i),g(T_j))$ ( $h(I_i,T_j)$ for short ), where the corresponding similarity score is defined as $S_{ij}=\sigma(F_{ij})$. 
Here, $\sigma(\cdot)$ denotes the sigmoid activation function.


\subsection{Energy-Guided Sample Filtration}
Our objective is to circumvent the pitfalls of previous methods that overconfidently divide samples with similarity prediction, thereby introducing potential sample selection risk.
Take the similarity scores [0.85, 0.80, 0.82] as an example: the first score represents the given sample pair, while the others correspond to its negative samples. Even though the given sample pair exhibits a high similarity score, it is not significantly different from the negative samples, suggesting a possible mismatch. Hence, selecting such a sample pair as ``clean'' based solely on similarity can be risky.

To this issue, by considering the overall prediction distribution, we aim to explore sample selection uncertainty to complement similarity-based sample filtration.
Given a  batchsize $B$, we first generate the classification logits $F_{i}$ of the visual input $I_i$ by viewing sample matching as a classification task within the batch. $F_{i}$ is formulated as $F_{i}=\{F_{i1},\cdots,F_{iB}\}$ with a corresponding label $y_i=i$.
Due to DNN's memorization effect, the model initially becomes adept at recognizing clean samples, leading to an unimodal distribution at $y_i$. 
In contrast, model struggles to differentiate noisy correspondences from their negatives, also giving rise to a more uniform distribution.

In view of such difference, we turn to energy uncertainty in logits space, which is a widely acceptable metric in the literature of uncertainty learning \cite{liu2020energy,xie2022active}.
Specifically, the energy uncertainty corresponding to the visual input $I_i$  can be calculated by:
\begin{align} 
    \operatorname{Energy}\left(I_i\right)=-\log {\textstyle{\sum_{b=1}^{B}} } e^{F_{ib}}.
\end{align}  
Intuitively, more uniformly distributed prediction (\ie, noisy correspondence) leads to higher estimated energy uncertainty \cite{zhang2023provable}.
Therefore we select the clean samples by applying a threshold $\tau$ and the maximum similarity constraint \cite{qin2022deep}, \ie,
\begin{equation}
  \mathcal{D}_{\text{clean}} = \{i\mid\text{Energy}(I_i)<\tau\ \text{and} \ \ y_i=\arg\textstyle{\max_j}F_{ij}\},
  \label{eq: filter}
\end{equation}
while  $\mathcal{D}_{\text{noisy}}$ refers to mismatched samples. In this sense, the selected samples maintain both low uncertainty and high similarity, paving the way for more precise sample division.

Moreover, we conceive an energy-bounded loss $L_u^I$ to reduce the energy uncertainty of clean samples while enhancing that of noisy samples, enlarging the margin between matched and mismatched samples, \ie,
\begin{equation}
  \begin{aligned}
    L_{u}^I & =\mathbb{E}_{i\sim \mathcal{D}_{\text {clean }}}[0, \text{Energy}(I_i)-m_{\text {clean}}]_+^{2} \\
    & +\mathbb{E}_{i\sim \mathcal{D}_{\text {noisy }}}[0, m_{\text {noisy}}-\text{Energy}(I_i)]_+^{2},
    \end{aligned}
\end{equation}
where $[x]_+=\max(x,0)$; $m_{\text {clean}}$ and $m_{\text {noisy}}$ are separate margins that penalize the clean (noisy) samples with energy uncertainty higher (lower) than the given margin.


\subsection{Swapped Gradient Weighting}
After sample filtration, it is risky to directly train the model on $\mathcal{D}_{\text{clean}}$ as it potentially contains some false positives \cite{huang2021learning,yang2023bicro}.
To ensure robust training, it's crucial to devise strategies that allow the model to adaptively maintain varied sensitivities to samples within $\mathcal{D}_{\text{clean}}$.
Instead of overconfident single similarity score, we introduce classification entropy to estimate sensitivity of each clean sample.
Visual input  $I_i$'s classification distribution is defined as $P_i=\text{softmax}(F_i)$, and the corresponding normalized classification entropy $e(P_i)$ is formulated as:
\begin{equation}
  e(P_i)=-\frac{\sum_{j=1}^{B}\left(P_{ij} \log P_{ij}\right)}{\log B}.
\end{equation}
Here $\log B$ is the maximum entropy to scale $e(P_i)$ into $[0,1]$ for numerical stability.
In this sense, low $e(P_i)$ highlights the model's ability to recognize matched samples, while suppressing similarity scores to other negative samples.
Consequently, model should be more sensitive to samples with lower $e(P_i)$ in optimization \cite{iscen2019label}.

In light of above, let $w_i^{I}$ denote the entropy-based model's sensitivity to visual input $I_i$ in i2t retrieval, formulated by:
\begin{equation}
  w_i^{I} = 1 - e(P_i)\mathds{1}(\alpha-S_{ii}+\sigma(h(I_i,T_{\phi(i)}))),
\end{equation}
where $\alpha>0$ is the expected margin between positive and negative match; $\phi(i)=\arg\max_{j\neq i}(F_{ij})$ and $T_{\phi(i)}$ is the hard negative text of $I_i$, \ie, the negative text most similar to $I_i$ within the batch. 
Moreover, we employ indicator function $\mathds{1}(\cdot)$ to evaluate whether a sample and its hard negative have expected discrimination $\alpha$. This design avoids unnecessary gradients on samples exhibiting satisfactory discrimination, reducing the risk of overfitting.
Besides, we further employ swapped prediction strategy on calculated $e(P_i)$, which is widely used in cross-modal tasks for improving robustness \cite{andonian2022robust}.
Its key idea is to use the weights derived from one modality for the other modality, promoting cross-modal consistency in the learning process.
For example, we use $w_i^{T}$ derived from t2i classification entropy for i2t retrieval training, and vice-versa.
Specifically, we apply $w_i^{T}$ with  hinge-based ranking loss, defined as:
\begin{equation}
L_{w}^{i2t}=\mathbb{E}_{i\sim\mathcal{D}_{\text{clean}}}[\alpha-w_i^{T}S_{ii}+\sigma(h(I_i,T_{\phi(i)}))]_{+}.
\label{eq: ranking loss}
\end{equation}
As a result, the derivative of $L_w^{i2t}$ with respect to model parameters $\theta$ is given by the chain rule $\frac{\partial L_{w}^{i2t}}{\partial \theta}=\frac{\partial L_{w}^{i2t}}{\partial S}\frac{\partial S}{\partial \theta}$ with
\begin{align}
  -\frac{\partial L_{w}^{i2t}}{\partial S_{ij}}=\left\{\begin{matrix}
    w_i^{T}, &j=i\\
    -1, &j = \phi(i)\\
    0,&otherwise
    \end{matrix}\right..
\label{derivation} 
\end{align}
\Cref{derivation} implies that clean samples exhibiting more certain distributions will retain larger gradients, consequently to which model is more sensitive.
Compared to sample-reweighting methods \cite{wei2021universal,wang2019multi}, our SGW strategy further suppresses similarity scores to hard negatives as $-\frac{\partial L_{w}^{i2t}}{\partial S_{i\phi(i)}}=-1$. 
Thus, \Cref{eq: ranking loss} can effectively adjust model's sensitivity of different samples in optimization, enhancing matching robustness.

\begin{table*}[!ht]
 \setlength{\abovecaptionskip}{0.cm}
\setlength{\belowcaptionskip}{0.cm}
	\newcommand{\tabincell}[2]{\begin{tabular}{@{}#1@{}}#2\end{tabular}}
	\centering
	\caption { Image-Text Retrieval on Flickr30K and MS-COCO 1K. Results marked with `*' are reproduced results from their official code, while `\dag' signifies methods that incorporate additional priors.}
	\resizebox{\textwidth}{!}{ 
		\begin{tabular}{c|c|ccc|ccc|c|ccc|ccc|c}
		\toprule[1.5pt]
		&&\multicolumn{7}{c|}{Flickr30K}&\multicolumn{7}{c}{MS-COCO}\\
		&&\multicolumn{3}{c|}{Image$\longrightarrow$Text}&\multicolumn{3}{c|}{Text$\longrightarrow$Image}&&\multicolumn{3}{c|}{Image$\longrightarrow$Text}&\multicolumn{3}{c|}{Text$\longrightarrow$Image}&\\
		\hline
			Noise&Methods&R@1&R@5&R@10&R@1&R@5&R@10&Sum&R@1&R@5&R@10&R@1&R@5&R@10&Sum\\
			\midrule
			\multirow{9}{*}{20\%}&SCAN& 58.5   & 81.0  & 90.8 & 35.5 & 65.0  & 75.2 & 406.0  & 62.2 & 90.0  & 96.1 & 46.2 & 80.8 & 89.2 & 464.5\\
			~&VSRN& 33.4   & 59.5 & 71.3 & 25.0  & 47.6 & 58.6 & 295.4 & 61.8 & 87.3 & 92.9 & 50.0  & 80.3 & 88.3 & 460.6 \\
			~&IMRAM & 22.7   & 54.0  & 67.8 & 16.6 & 41.8 & 54.1 & 257.0  & 69.9 & 93.6 & 97.4 & 55.9 & 84.4 & 89.6 & 490.8 \\
			~&SAF & 62.8   & 88.7 & 93.9 & 49.7 & 73.6 & 78.0  & 446.7 & 71.5 & 94.0  & 97.5 & 57.8 & 86.4 & 91.9 & 499.1 \\
			~&SGR& 55.9   & 81.5 & 88.9 & 40.2 & 66.8 & 75.3 & 408.6 & 25.7 & 58.8 & 75.1 & 23.5 & 58.9 & 75.1 & 317.1 \\
			~&NCR & 73.5   & 93.2 & 96.6 & 56.9 & 82.4 & 88.5 & 491.1 & 76.6 & 95.6 & 98.2 & 60.8 & 88.8 & 95.0  & 515.0  \\
			~&DECL&77.5&93.8 &97.0&56.1&81.8&88.5&494.7&77.5&95.9&98.4&61.7&89.3&95.4&518.2\\
			~&BiCro\dag&78.1&94.4&97.5&60.4&84.4&89.9&504.7&78.8&96.1&98.6&63.7&90.3&95.7&523.2\\
			~&MSCN\dag&77.4&\textbf{94.9}&97.6&59.6&83.2&89.2&502.1&78.1&\textbf{97.2}&98.8&\textbf{64.3}&90.4&95.8&\textbf{524.6}\\
			~&RCL&75.9&94.5&97.3&57.9&82.6&88.6&496.8&\textbf{78.9}&96.0&98.4&62.8&89.9&95.4&521.4\\
			~&\textbf{Ours}&\textbf{79.5}&94.2&\textbf{97.9}&\textbf{61.2}&\textbf{84.8}&\textbf{90.2}&\textbf{507.8}&78.5 & 96.8 & \textbf{98.8} & 63.8 & \textbf{90.4 }& \textbf{95.8} & 524.1 \\
			\midrule
			\multirow{9}{*}{40\%}&SCAN& 26.0    & 57.4 & 71.8 & 17.8 & 40.5 & 51.4 & 264.9 & 42.9 & 74.6 & 85.1 & 24.2 & 52.6 & 63.8 & 343.2 \\
			~&VSRN& 2.6   & 10.3 & 14.8 & 3.0  & 9.3 & 15.0  & 55.0  & 29.8 & 62.1 & 76.6 & 17.1 & 46.1 & 60.3 & 292.0  \\
			~&IMRAM& 5.3   & 25.4 & 37.6 & 5.0  & 13.5 & 19.6 & 106.4 & 51.8 & 82.4 & 90.9 & 38.4 & 70.3 & 78.9 & 412.7 \\
			~&SAF & 7.4   & 19.6 & 26.7 & 4.4 & 12.2 & 17.0  & 87.3 & 13.5 & 43.8 & 48.2 & 16.0  & 39.0  & 50.8 & 211.3 \\
			~&SGR& 4.1   & 16.6 & 24.1 & 4.1 & 13.2 & 19.7 & 81.8 & 1.3 & 3.7 & 6.3 & 0.5 & 2.5 & 4.1 & 18.4 \\
			~&NCR & 68.1   & 89.6 & 94.8 & 51.4 & 78.4 & 84.8 & 467.1 & 74.7 & 94.6 & 98.0  & 59.6 & 88.1 & 94.7 & 509.7 \\
			~&DECL&72.7&92.3&95.4&53.4&79.4&86.4&479.6&75.6&95.5&98.3&59.5&88.3&94.8&512.0\\
			~&BiCro\dag&74.6&92.7&96.2&55.5&81.1&87.4&487.5&77.0&95.9&98.3&61.8&89.2&94.9&517.1\\
			~&MSCN*\dag&71.9&92.0&95.4&55.1&80.2&86.8&481.3&77.1&95.7&98.4&61.2&88.6&94.8&515.7\\
			~&RCL&72.7&92.7&96.1&54.8&80.0&87.1&483.4&77.0&95.5&98.3&61.2&88.5&94.8&515.3\\
			~&\textbf{Ours}&\textbf{76.5}&\textbf{93.9}&\textbf{96.3}&\textbf{57.5}&\textbf{82.7}&\textbf{88.5}&\textbf{495.4}& \textbf{77.2} &\textbf{96.0} &\textbf{98.5} &\textbf{62.1} &\textbf{89.3} &\textbf{95.3} &\textbf{518.4} \\
			\midrule
			\multirow{9}{*}{60\%}&SCAN& 13.6     & 36.5 & 50.3 & 4.8  & 13.6 & 19.8 & 138.6 & 29.9 & 60.9 & 74.8 & 0.9  & 2.4  & 4.1  & 173.0   \\
			~&VSRN & 0.8      & 2.5  & 5.3  & 1.2  & 4.2  & 6.9  & 20.9  & 11.6 & 34.0   & 47.5 & 4.6  & 16.4 & 25.9 & 140.0   \\
			~&IMRAM & 1.5      & 8.9  & 17.4 & 1.9  & 5.0    & 7.8  & 42.5  & 18.2 & 51.6 & 68.0   & 17.9 & 43.6 & 54.6 & 253.9 \\
			~&SAF  & 0.1      & 1.5  & 2.8  & 0.4  & 1.2  & 2.3  & 8.3   & 0.1  & 0.5  & 0.7  & 0.8  & 3.5  & 6.3  & 11.9  \\
			~&SGR & 1.5      & 6.6  & 9.6  & 0.3  & 2.3  & 4.2  & 24.5  & 0.1  & 0.6  & 1.0    & 0.1  & 0.5  & 1.1  & 3.4   \\
			~&NCR  & 13.9     & 37.7 & 50.5 & 11.0   & 30.1 & 41.4 & 184.6 & 0.1  & 0.3  & 0.4  & 0.1  & 0.5  & 1.0    & 2.4   \\
			~&DECL&65.2&88.4&94.0&46.8&74.0&82.2&450.6&73.0&94.2&97.9&57.0&86.6&93.8&502.5\\
			~&BiCro\dag&67.6&90.8&94.4&51.2&77.6&84.7&466.3&73.9&94.4&97.8&58.3&87.2&93.9&505.5\\
			~&MSCN*\dag&67.5&88.4&93.1&48.7&76.1&82.3&456.1&74.1&94.4&97.6&57.5&86.4& 93.4& 503.4\\
			~&RCL&67.7&89.1&93.6&48.0&74.9&83.3&456.6&74.0&94.3&97.5&57.6&86.4&93.5&503.3\\
			~&\textbf{Ours}&\textbf{71.0}&\textbf{92.1}&\textbf{96.1}&\textbf{54.0}&\textbf{80.1}&\textbf{87.0}&\textbf{480.3}&\textbf{74.5}&\textbf{94.5}&\textbf{97.9}&\textbf{58.7}&\textbf{87.5}&\textbf{93.9}&\textbf{506.9}\\
			
			
			\bottomrule[1.5pt]
	\end{tabular}}
	\label{table:flicker}
\end{table*}
\setlength{\belowcaptionskip}{-0mm}

\subsection{Cross-Modal Biased Complementary Learning}
Evidently, \Cref{derivation} highlights that $L_w^{i2t}$ overlooks numerous negative similarities defined as:
\begin{equation}    
\{S_{ij}\mid j\neq i; \text{ and if } i \in \mathcal{D}_{\text{clean}}, j \neq \phi(i)\}.
\end{equation}
These overlooked negative similarities maintain zero gradients and are ignored in model optimization.
However, in classification, these overlooked similarities indicate the samples that do not match the given sample, \ie, complementary labels.
As shown in \Cref{fig: overview}, harnessing these complementary labels can enhance the stability of the model optimization.
In this sense, we construct an auxiliary dataset $\mathcal{D}_{neg}=\{(i, \bar{\mathcal{Y}}_i)\}_{i=1}^{B}$ within each batch. 
Here $i$ is the index of given image $I_i$ within batch, $\bar{\mathcal{Y}}_i$ is corresponding complementary labels formulated as:
\begin{equation}
\bar{\mathcal{Y}}_i=\{0, 1,\cdots,B-1\} \setminus \{y_i \}.
\end{equation}
Furthermore, we explore non-uniformly distributed complementary labels to improve model's generality, due to the following facts:
1) Ideal uniformly distributed complementary labels do not necessarily hold in real-world data, particularly in instance-level classification.
2) Non-uniform complementary labels permit model to focus more on the harder negatives, thereby preventing informative supervision from being overwhelmed by redundant negative samples.

Inspired by \cite{yu2018learning,gao2021discriminative}, we prefer to choose negative texts that have higher similarity to $I_i$ as the complementary label, enabling model to focus more on challenging and informative negative counterparts. 
Notably, we directly use similarity to estimate the selection probability of complementary labels, due to the fact that complementary labels are leveraged to suppress negative information and do not involve self-reinforcing errors.
Specifically, we employ MS \cite{wang2019multi} to gauge the likelihood of selecting $T_j$ as $I_i$'s complementary label. MS considers both self and relative similarities, defined as:
\begin{equation}
  \bar{P}_{ij}^{i2t}= \frac{e^{\beta\left(S_{i j}-b\right)}}{1+\sum_{k \in \bar{\mathcal{Y}}_i} e^{\beta\left(S_{i k}-b\right)}},
    \label{eq: reweight}
\end{equation}
where $\beta$ and $b$ are two hyperparameters of Binomial deviance \cite{hastie2009elements}, controlling the smoothness of selection distribution.
Note that selected hard negatives from $\mathcal{D}_{\text{clean}}$ have already been considered in $L_w^{i2t}$, we exclude these samples to prevent their over-representation in the model training process, which is formulated by:
\begin{equation}
    \bar{P}_{i\phi(i)}^{i2t}=-\infty, \forall i\in\mathcal{D}_{\text{clean}}.
\end{equation}
We then rectify complementary labels using the overall selection probability \cite{yu2018learning}, \ie,
\begin{equation}
  S^\prime = \text{softmax}(\bar{P}^{i2t})^TS.
\end{equation}
Ultimately, the cross-modal biased complementary learning objective $L_c^{i2t}$ on $\mathcal{D}_{neg}$ is formulated as:
\begin{equation}
  L_c^{i2t}=-\mathbb{E}_{(i,\bar{\mathcal{Y}}_i)\sim \mathcal{D}_{neg}}\mathbb{E}_{j\sim\bar{\mathcal{Y}}_i}[\log(1-S^\prime_{ij})].
  \label{eq: cll}
\end{equation}

\paragraph{Theoretical Analyses}
We provide theoretical evidence to better elucidate the effectiveness of our CMBCL.
Specifically, we demonstrate CMBCL's efficacy in \Cref{optimal} as:
\begin{theorem}
  Given suﬃcient data with complementary labels, minimizing \Cref{eq: cll} can yield the optimal classifier equivalent to that trained with the true labels.
\label{optimal}
\end{theorem}
Additionally, CMBCL, by considering complementary labels' distribution, exhibits superior generalizability. 
In detail, CMBCL generalizes previous strong competitor \cite{hu2023cross} as a special case.

\subsection{Model Optimization}
To ensure consistent performance across modalities, we employ SREM for bidirectional matching, encompassing both image-to-text and text-to-image tasks, formulated by:
\begin{equation*}
    \min_{\theta} L=0.5(L_w^{i2t}+L_w^{t2i})+\lambda_1(L_u^I+L_u^T)+\lambda_2(L_c^{i2t}+L_c^{t2i}),
\end{equation*}
where $L_u^T$, $L_w^{t2i}$, and $L_c^{t2i}$ represent objectives when symmetrically applying energy-guided sample filtration, SGW, and CMBCL for text-to-image retrieval.
$\lambda_1, \lambda_2\in[0,1]$ are hyperparameters to adjust the effect of energy uncertainty estimation and negative information utilization.
\section{Experiments}
\subsection{Experiments Setting}
\paragraph{Datasets}
Following previous works, we evaluate our SREM using three image-text retrieval datasets, including COCO \cite{lin2014microsoft}, Flickr30K \cite{young2014image} and CC152K \cite{huang2021learning}. The first two are well-annotated, while the third one is harvested from the internet. We provide an overview of the dataset details as follows:
\begin{itemize}
  \item COCO and Flickr30K contain 123287 and 31783 images with 5 corresponding captions per image, respectively. Following \cite{huang2021learning}, we maintain 5K/5K and 5K/5K image-text pairs for validation/test, leaving the remainder for training.
  \item CC152K is a subset of Conceptual Captions \cite{sharma2018conceptual} containing 152K image-text pairs. We use 150K pairs for training, 1K for validation and another 1K for testing.
\end{itemize}

\paragraph{Evaluation Metrics}
Following previous work \cite{han2023noisy}, we evaluate SREM with the recall rate at K (R@K) that measures the proportion of relevant items found within the top K results of a ranked list. By querying both images and texts, we report corresponding results of R@1, R@5 and R@10, which are further summed to evaluate the overall performance, \ie, R\_sum.

\paragraph{Implementation Details}
As a plug-and-play module, our SREM can be seamlessly applied in various image-text retrieval methods to improve their robustness against noisy correspondences. Here, we adopt the same backbone, SGRAF \cite{diao2021similarity}, with the same training settings as \cite{huang2021learning} for fair comparisons.
Specifically, we warm up the model for 5 epochs with $L_c^{i2t}$ and $L_c^{t2i}$ to achieve initial convergence, followed by a 50 epochs training process.
We employ a batch size of 128 and an Adam \cite{kingma2014adam} optimizer with a learning rate of $2e$-4 that will be decayed by $0.1$ after 25 epochs.

\begin{figure*}[!t]
    \setlength{\belowcaptionskip}{-5mm}
    \setlength{\abovecaptionskip}{1mm}
    \centering
    \subfloat[Initial distribution]{{\includegraphics[width=0.24\textwidth]{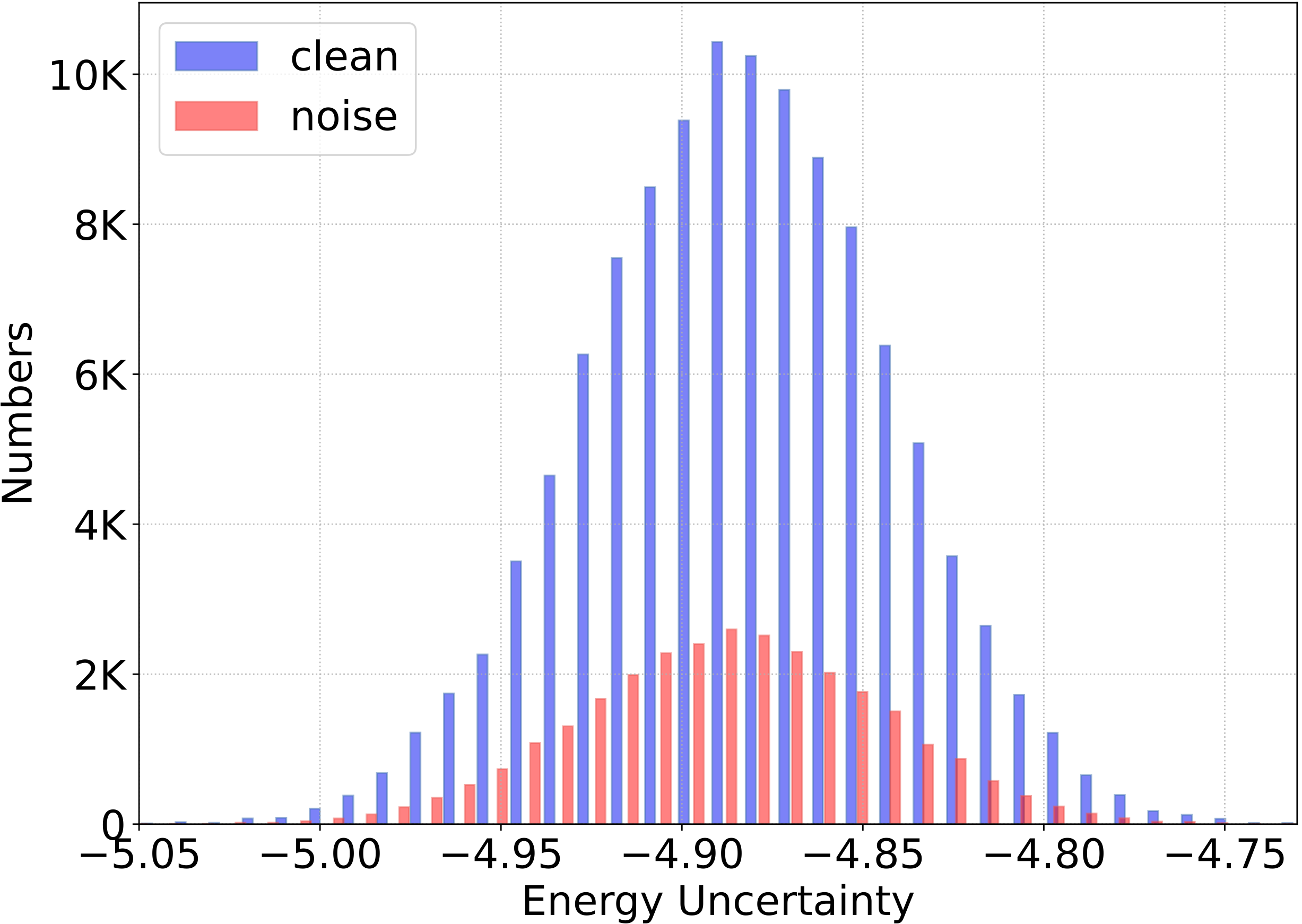}}} \hfill
    \subfloat[Epoch 10]{{\includegraphics[width=0.24\linewidth]{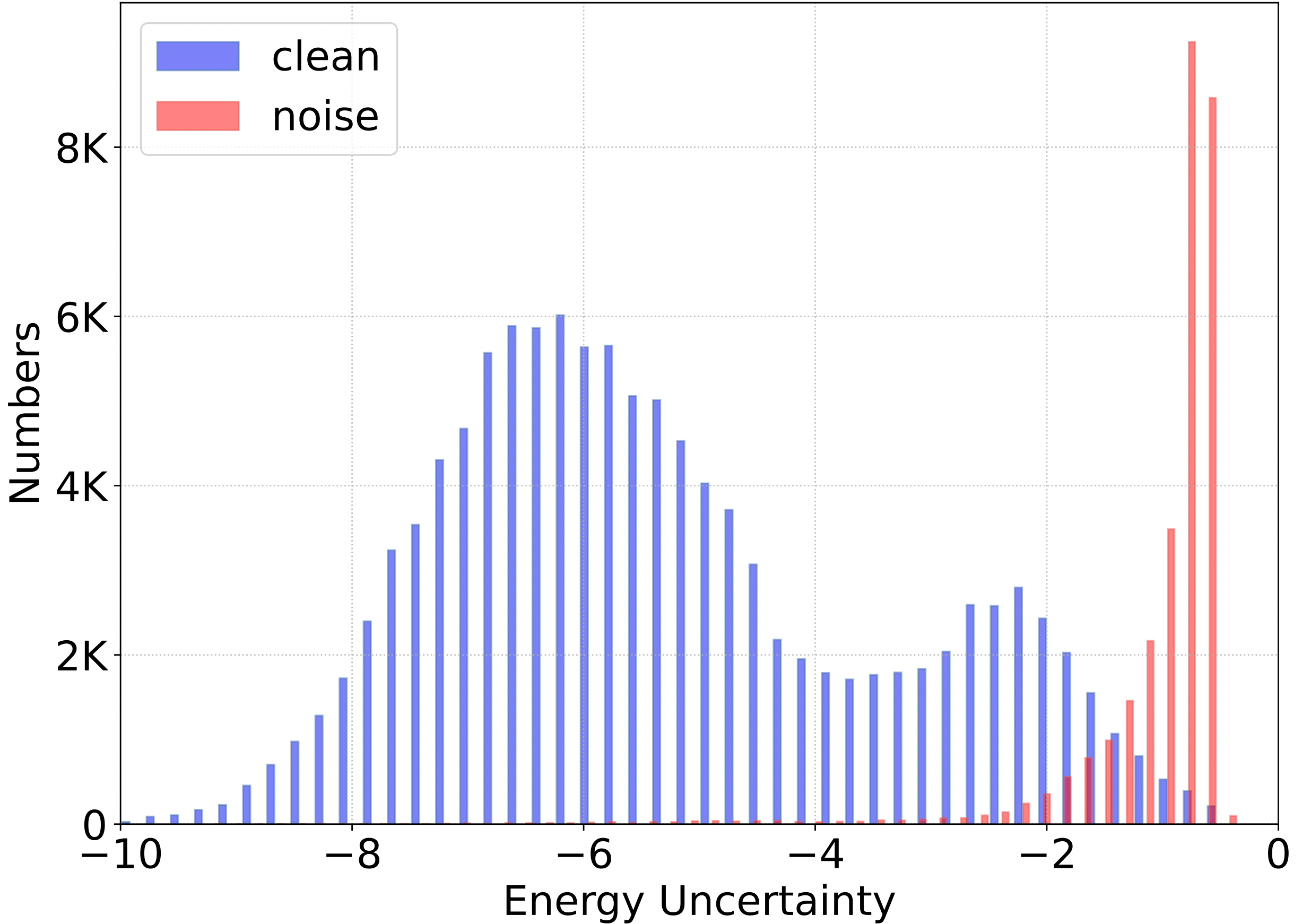}}} \hfill
    \subfloat[Epoch 30]{{\includegraphics[width=0.24\linewidth]{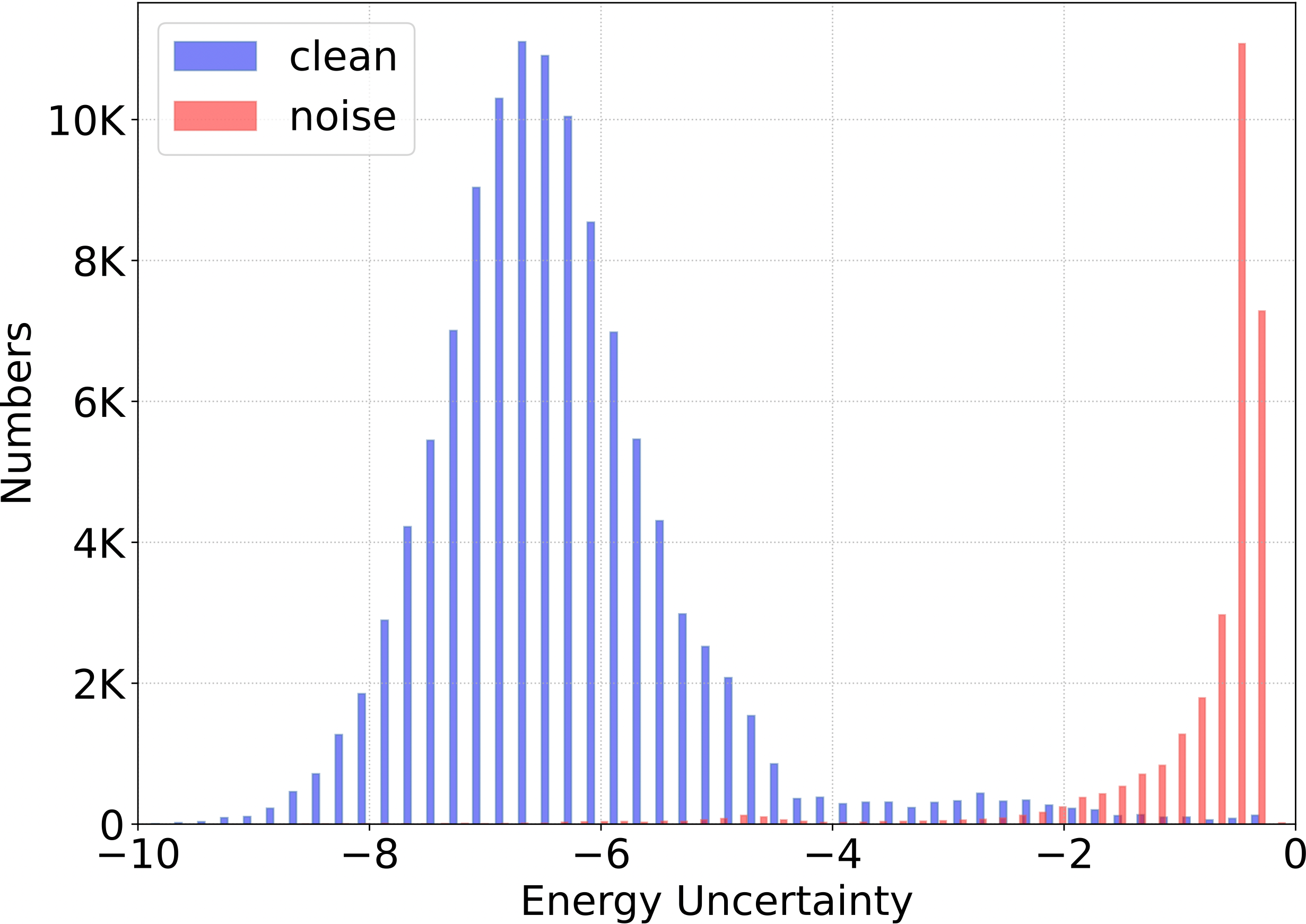}}} \hfill
    \subfloat[Epoch 50]{{\includegraphics[width=0.24\linewidth]{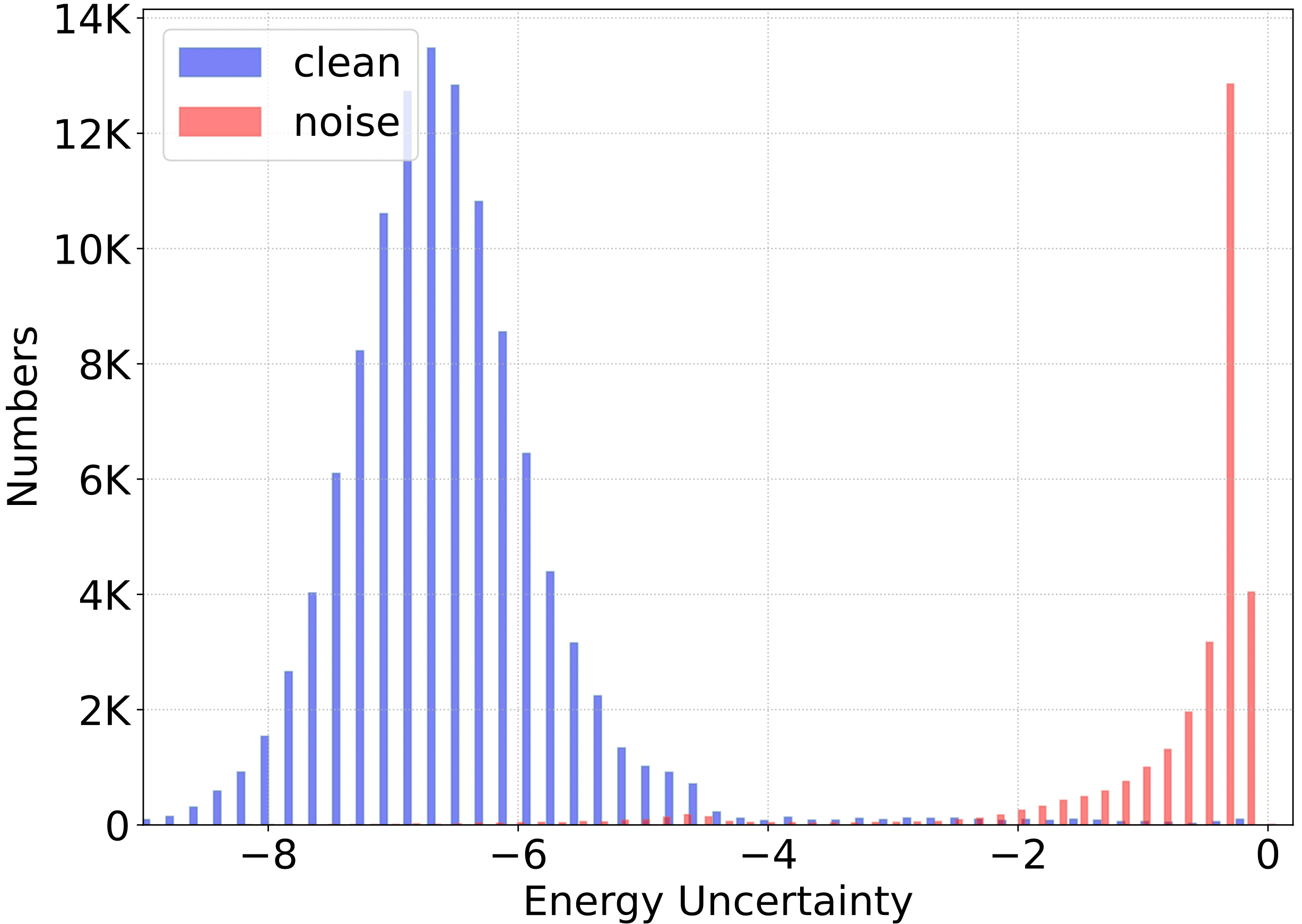}}}
  \caption{We visualize the energy uncertainty distribution of clean and noisy pairs at different training stages of our SREM, which is conducted on Flickr30K under 20\% noise. Thanks to SREM, the energy uncertainty of clean pairs gradually approaches the left (low) and the energy uncertainty of noisy pairs tightly gathers to the right (high).}
  \label{fig: uncertainty}
  \end{figure*}
\subsection{Comparison with State-Of-The-Art}
We compare the proposed SREM against current state-of-the-art (SOTA) methods to demonstrate its effectiveness, including general image-text retrieval methods SCAN \cite{lee2018stacked}, VSRN \cite{li2019visual}, IMRAM \cite{chen2020imram}, SGR, SAF\cite{diao2021similarity}, and noisy correspondence robust methods NCR \cite{huang2021learning}, DECL \cite{qin2022deep}, MSCN \cite{han2023noisy}, BiCro \cite{yang2023bicro} and RCL \cite{hu2023cross}.

\begin{table}[t!]
	\setlength{\abovecaptionskip}{0.cm}
    \setlength{\belowcaptionskip}{0.cm}
	\makeatletter\def\@captype{table}
	\caption{Image-Text Retrieval on CC152K.}
	\resizebox{\linewidth}{!}{
		\begin{tabular}{c|ccc|ccc|c}
			\toprule[1.5pt]
			& \multicolumn{3}{c|}{Image$\longrightarrow$Text}&\multicolumn{3}{c|}{Text$\longrightarrow$Image}&\\
			\hline
			Methods&R@1&R@5&R@10&R@1&R@5&R@10&Sum\\
			\hline
			SCAN&30.5&55.3&65.3&26.9&53.0&64.7&295.7\\
			VSRN&32.6&61.3&70.5&32.5&59.4&70.4&326.7\\
			IMRAM&33.1&57.6&68.1&29.0&56.8&67.4&312.0\\
			SAF&31.7&59.3&68.2&31.9&59.0&67.9&318.0\\
			SGR&11.3&29.7&39.6&13.1&30.1&41.6&165.4\\
			NCR&39.5&64.5&73.5&40.3&64.6&73.2&355.6\\
			DECL&39.0&66.1&75.5&40.7&66.3&76.7&364.3\\
			BiCro&40.8&67.2&76.1&\textbf{42.1}&67.6&76.4&370.2\\
			MSCN&40.1&65.7&76.6&40.6&67.4&76.3&366.7\\
			\textbf{Ours}&\textbf{40.9}&\textbf{67.5}&\textbf{77.1}&41.5&\textbf{68.2}&\textbf{77.0}&\textbf{372.2}\\
			
			\bottomrule[1.5pt]
		\end{tabular}
	}
  \label{cc152k}
  \end{table}
\subsubsection{Results on Synthetic Noise of Flickr30K and MS-COCO}
As in previous works, we emulate noisy correspondences by randomly shuffling the training images and captions for specific noise ratios. 
We report results with noise ratio $20\%$, $40\%$, $60\%$ for comprehensive comparison with current SOTAs, such as MSCN and BiCro. 

\Cref{table:flicker} details the results of Flickr30K and MS-COCO on different noise ratios, where the results of MS-COCO are averaged on 5 folds of 1K test images as in previous works. 
We find that the strong noise-robust competitors, \ie, MSCN and BiCro, achieve markedly better results than general image-text retrieval methods, highlighting the necessity of designing models that can effectively withstand noise.
However, they introduce strong priors, \eg, $3\%$ additional clean samples for MSCN and extra model ensemble for BiCro, resulting in costly data collection and computation overhead, respectively.
More troublingly, as the noise ratio increases, the performance of these methods deteriorates drastically due to self-reinforcing errors.
In contrast, our SREM, devoid of any such priors, is more effective and stable, improving R\_sum by more than 1\% on average. 

\subsubsection{Results on Real-World Noise of CC152K}
CC152K, automatically harvested from the Internet, inherently contains approximately $20\%$ noisy correspondences.
It thereby can be used to evaluate SREM's ability in handling real-world noise. 
We train and evaluate SREM without introducing any additional synthetic noise.
\Cref{cc152k} shows that SREM performs commendably even without any priors.
Specifically, it outperforms the strongest competitors MSCN and BiCro by an average of 1\% in R\_sum.
Besides, SREM consistently and significantly triumphs over all baselines in all results, except for R@1 of retrieving images.
These results demonstrate SREM's appealing efficacy in real-world scenarios.


        

\begin{table}[t!]
  \centering
  \scriptsize
  \setlength{\tabcolsep}{3.1pt}
    \setlength{\abovecaptionskip}{0.cm}
    \setlength{\belowcaptionskip}{0.cm}
    \makeatletter\def\@captype{table}
    \caption{Ablation studies on CC152K with real-world noise.}
    \resizebox{\linewidth}{!}
    {
      \begin{tabular}{cccc|c|c}
        \toprule[1.5pt]
        \multicolumn{4}{c|}{Methods}& \multicolumn{1}{c|}{Image$\longrightarrow$Text}&\multicolumn{1}{c}{Text$\longrightarrow$Image}\\
        \hline
        \thead{Sample \\ Filtration}&\thead{Gradient \\ Weighting}&\thead{Complementary \\ Learning}&\thead{Label \\ Rectification}&R@1/5/10&R@1/5/10\\
        \bottomrule[0.5pt]
        & & & &32.5/59.5/70.0&32.5/60.7/68.7\\
        \checkmark&& & &37.3/63.7/73.1&36.9/64.8/74.1\\
        \checkmark&\checkmark& & &40.2/63.2/74.2&37.7/65.3/74.9\\
        \checkmark&\checkmark&\checkmark&&40.5/67.3/75.8&\textbf{42.5}/67.9/76.2\\
        \checkmark&\checkmark&\checkmark&\checkmark&\textbf{40.9}/\textbf{67.5}/\textbf{77.1}&41.5/\textbf{68.2}/\textbf{77.0}\\
        
        \bottomrule[1.5pt]
      \end{tabular}
    }   
    \label{table:ablation}
  \end{table}
  
  \begin{figure*}[!t]
    \setlength{\belowcaptionskip}{-4mm}
    \setlength{\abovecaptionskip}{1mm}
    \centering
    \includegraphics[width=.99\linewidth]{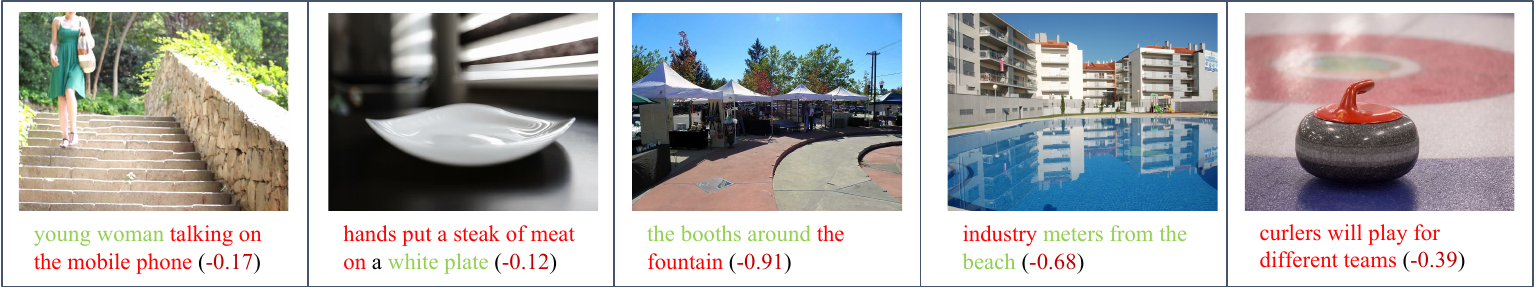}
  \caption{Real-world noisy examples detected by our SREM, with the setting of $m_{clean}=-4$ and $m_{noisy}=0$, whose energy uncertainty are shown in brackets. We highlight the matched words in green and the mismatched words in red.}
  \label{fig: real-word}
  \end{figure*}

\subsection{Ablation Studies}
This section conducts ablation studies to comprehensively evaluate the performance of each component in SREM.
\subsubsection{Component Analyses}
\Cref{table:ablation} shows that vanilla trained model exhibits suboptimal performance, illustrating its susceptibility to disturbances caused by noisy correspondences.
The energy-guided sample filtration significantly enhances the performance by more than 10\% on R@1.
When using swapped gradient weighting, we observe performance boosts in all results, except R@5 for text retrieving.
Furthermore, leveraging unused negative information as complementary labels considerably improves performance, evidenced by an increase of more than 1\% in R\_sum.
The rectification of complementary labels further enhances performance, validating the efficacy of considering complementary label distributions.
These results underline the significant role of complementary labels in fortifying retrieval robustness.
The best performance is achieved with all proposed components, demonstrating their efficacy. 
\subsubsection{Visualization on Energy Uncertainty}
\Cref{fig: uncertainty} visualizes energy uncertainty during training.
As training progresses, the energy uncertainty of clean samples becomes lower while that of noisy correspondences increases, manifesting a clear polarizing trend.
These observations validate the efficacy of energy uncertainty estimation for noisy correspondences.
Therefore, the energy uncertainty from the overall prediction distribution can naturally be used to differentiate between noisy and clean pairs, further boosting the robustness against noisy correspondences.
\subsubsection{Visualization on Self-Reinforcing Errors}
\begin{figure}[!t]
\setlength{\abovecaptionskip}{1mm}
\setlength{\belowcaptionskip}{-5mm}
  \centering
  \includegraphics[width=1\linewidth]{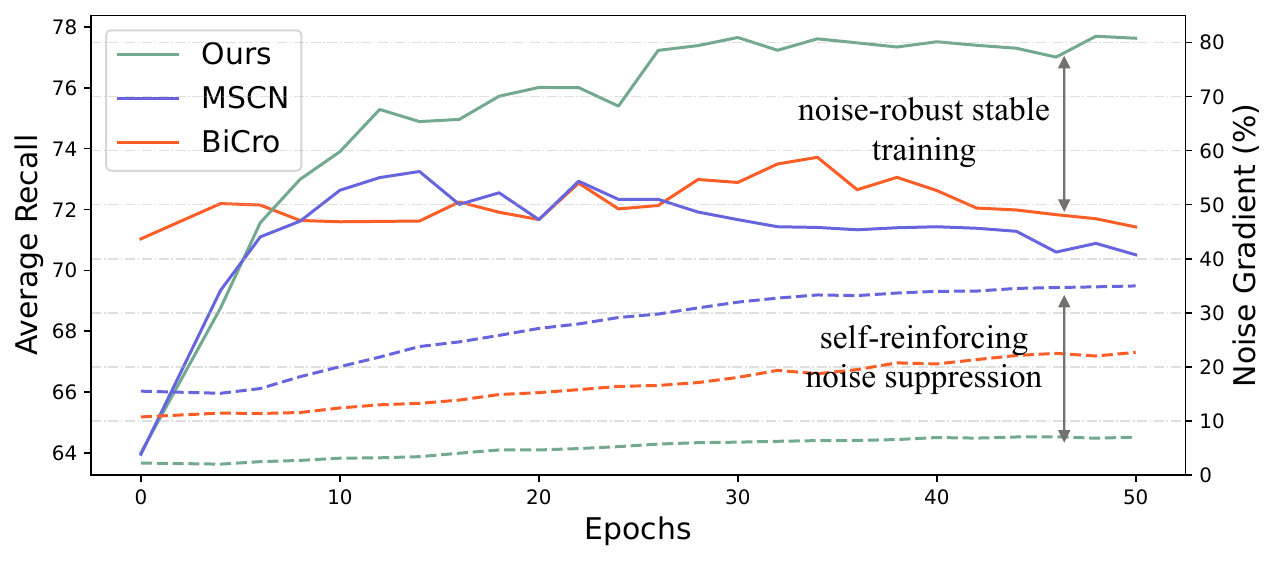}
\caption{Comparing performance (solid) and noise gradient ratio (dashed) on 60\% noise Flickr30K as training proceeds.}
\label{fig: gradient_rsum}
\end{figure}

We track the training progress to validate the efficacy of our SREM in alleviating self-reinforcing errors.
Specifically, we measure the performance of each epoch, as well as noisy gradients ratio relative to all positive gradients (the proportion of false positive gradients created by enhancing mismatched samples' similarity in $L_w$).
We also provide the results of MSCN and BiCro for more comprehensive and fair comparisons.
As shown in \Cref{fig: gradient_rsum}, since CMBCL during warmup avoids self-reinforcing errors, SREM starts with the lowest noisy gradient ratio. While in training, with its carefully designed components, SREM effectively suppresses self-reinforcing errors, exhibiting a significantly lower and stable noise gradient ratio, \ie, less than 7\%.
In contrast, MSCN and BiCro start with higher noise gradient ratios that rapidly increase in training due to their similarity-based training with hard negatives.
As a result, SREM achieves better performance with stable optimization, while MSCN and BiCro exhibit unsatisfactory results, whose performance gradually drops with noisy gradient ratio increasing.
These results highlight the efficacy of SREM in alleviating self-reinforcing errors.
\subsubsection{Efficiency Analyses}
It is noteworthy that SREM also maintains superior efficiency.
Specifically, \Cref{tab: efficiency} records the training overhead per epoch on CC152K using an NVIDIA Tesla A40 48G.
The training time of MSCN and BiCro contains two parts as they first pre-compute similarity across the entire dataset and then conduct sample filtration before training. These steps incur additional computation and storage overhead.
Moreover, MSCN computes meta gradients for model optimization and BiCro rectifies soft correspondences via numerous anchor samples, both of which are computationally expensive and thus further diminishing efficiency.
Differently, our SREM not only eliminates the pre-computation but also employs computationally efficient techniques, \ie, energy uncertainty, entropy and complementary learning.
Consequently, SREM reduces the training time by more than 40\%, highlighting its efficiency and potential applicability to large-scale datasets.

\subsubsection{Detected Noisy Correspondences in Real-World Scenario}
\Cref{fig: real-word} shows some real-world noisy correspondences in CC152K detected by our SREM with their corresponding energy uncertainty.
Specifically, SREM is not limited to recognizing only obvious noisy pairs containing completely irrelevant information. 
It also can identify hard mismatched pairs with subtle semantic misalignment, \eg, the missing elements of the phone, hands, steak and fountain, \etc, as well as the incongruence between concepts like ``building" and ``industry".
These results qualitatively demonstrate SREM's efficacy, revealing its promise for handling real-world applications.
\begin{table}[!t]
\setlength{\abovecaptionskip}{0.cm}
\setlength{\belowcaptionskip}{0.cm}
  \caption{Training efficiency comparison in terms of time cost and graphics memory. The reported per-epoch time is the average time for 50 epochs.}
  \label{tab: efficiency}
  \resizebox{\columnwidth}{!}{%
  \begin{tabular}{c|ccc}
  \toprule
  Methods & Pre-Filtration ($S$) & Training ($S$) & GPU Memory (MB) \\
  \midrule
  MSCN   & 365        & 6344    & 21367     \\
  BiCro  & 358        & 3093     & 14543     \\
  Ours   & 0          & 1506     & 13022    \\
  \bottomrule
  \end{tabular}%
  }
  \end{table}
\section{Conclusion}
This paper presents a novel framework, SREM, to address the challenges of noisy correspondences in cross-modal matching. Using per-sample classification logits, SREM ingeniously employs energy uncertainty to filter out the noisy correspondences, paving the way for more precise data division.
It then applies swapped classification entropy to recalibrate gradients, offering a more nuanced approach to assessing model's sensitivity in sample matching, compared to single similarity scores.
Moreover, the CMBCL framework within SREM harnesses previously overlooked negative information, ensuring stable model optimization.
Both theoretical evidence and extensive experiments on challenging benchmarks corroborate SREM's superiority in efficacy, efficiency and generality.
We hope our SREM will drive improvements in both the efficacy and efficiency of noisy correspondence learning, providing new insights into building more robust cross-modal information retrieval systems.

\section{Acknowledgement}
This work is supported by the National Key Research and Development Program of China (No. 2022YFB3102600), National Nature Science Foundation of China (No. 62192781, No. 62272374, No. 62202367, No. 62250009, No. 62137002), Project of China Knowledge Center for Engineering Science and Technology,  Project of Chinese academy of engineering ``The Online and Offline Mixed Educational Service System for `The Belt and Road' Training in MOOC China'', and the K. C. Wong Education Foundation.

\bibliography{aaai24}

\end{document}